\documentclass{sigkddExp}
\usepackage{cite}
\usepackage{graphicx}
\usepackage{algorithmic}
\usepackage{multicol}
\usepackage{multirow}
\usepackage{booktabs}
\usepackage{amsmath,amssymb,amsfonts}
\usepackage{microtype}
\usepackage{verbatim}
\usepackage{fancyvrb}
\usepackage{fvextra} 
\usepackage{inconsolata}
\usepackage{subfigure}
\usepackage{hyperref}
\usepackage{enumitem} 
\usepackage{siunitx} 
\begin{document}
\title{Urban Planning in the Age of Agentic AI: Emerging Paradigms and Prospects}

\numberofauthors{1}

\author{Rui Liu$^{1}$, Tao Zhe$^{1}$, Zhong-Ren Peng$^{2}$, \\ Necati Catbas$^{3}$, Xinyue Ye$^{4}$, Dongjie Wang$^{1,\dagger}$, Yanjie Fu$^{5,\dagger}$ \\ \\
\textsuperscript{1}University of Kansas,
\textsuperscript{2}University of Florida,
\textsuperscript{3}University of Central Florida,\\
\textsuperscript{4}University of Alabama,
\textsuperscript{5}Arizona State University\\
\small{$\dagger$ Corresponding authors: Dongjie Wang and Yanjie Fu.}
}












\maketitle

\begin{abstract}
Generative AI, large language models (LLMs), and agentic AI have emerged separately of urban planning.  However, the convergence between AI and urban planning presents an interesting opportunity towards AI urban planners. 
Existing studies conceptualizes urban planning as a generative AI task, where AI synthesizes land-use configurations under geospatial, social, and human-centric constraints and reshape automated urban design.  
We further identify critical gaps of existing generative urban planning studies: 1) the generative structure has to be predefined with strong assumption: all of adversarial networks, diffusion models, hierarchical zone-POI generative structure are predefined by humans; 2) ignore the power of domain expert developed tools: domain urban planners have developed various tools in the urban planning process guided by urban theory, while existing pure neural networks based generation ignore the power of the tools developed by urban planner practitioners. 
To address these limitations, we  outline a future research direction agentic urban AI planner, calling for a new synthesis of agentic AI and participatory urbanism that integrates AI capabilities with domain expertise and public engagement.
\end{abstract}

\section{Introduction and Motivation}
\vspace{12pt}
Urban planning is an interdisciplinary and complex process that involves public policy, social science, engineering, architecture, landscape, and other related fields. 
Effective urban planning can help to mitigate the operational and social vulnerability of a urban system, such as high tax, crimes, traffic congestion and accidents, pollution, depression, and anxiety~\cite{adams1994urban,goetz2020whiteness,yiftachel1989towards}. 

Traditional urban planning processes are mostly completed by professional human planners. 
Although urban data are increasingly available, the classic planning methods remain static and slow to adapt. 
However, rapid urbanization, climate change, aging infrastructure, and widening social inequities together create challenges on how cities are designed. 

1Recent breakthroughs in Generative AI (GenAI)~\cite{radford2015unsupervised,ho2020denoising, oussidi2018deep,noyman2020deep}, Large Language Models (LLMs)~\cite{achiam2023gpt, brown2020language,touvron2023llama}, and Agentic AI~\cite{durante2024agent,shavit2023practices} offer a transformative opportunity to rethink urban planning. 
These technologies introduce the abilities to generate alternative futures, understand complex textual and visual data, and act as autonomous agents that reason, simulate, and navigate planning goals. The new abilities of these AI paradigms make it possible to develop an AI Urban Planner that can augment human expertise, democratize access to planning insights, and adapt to changing urban conditions. 
This insight prompts a fundamental question: how might AI reshape the practice and ethics of urban planning in the era of AI? What roles does GenAI, LLMs, and agentic AI play in urban planning? Can machines develop and learn at a human capability to automatically and quickly calculate land-use configurations? In this way, machines can be planning assistants, while human planners can adjust machine-generated plans for specific needs.

\section{Existing Studies: Urban Planning as Generative AI}
\vspace{12pt}
Although urban planning is complex and difficult to automate, a new perspective is to formulate urban planning as a generative task that generates optimal land uses and built environment configurations conditional on geospatial, infrastructural, socioeconomic, and human-centric constraints specific to targeted regions.

We propose to leverage and adapt these GenAI models to produce diverse land-use configurations and generate spatial designs conditioned on multimodal inputs, such as urban geography data, human mobility patterns, and environmental constraints~\cite{wang2023towards}. 
One example task is to create alternative zoning proposals that meet goals for density, connectivity, or climate resilience, while incorporating learned priors from historical planning outcomes. 
Such new capabilities enable urban planners to shift from rule-based prescriptions to data-driven generative design that is both adaptive and customizable~\cite{yuan2014discovering}.

\subsection{Recent Studies in GenAI for Urban Planning}
\vspace{12pt}
Drawing on the advanced capabilities of generative AI, we have reenvisioned urban planning as a generative AI task, conceptualizing spatial configuration plans as images, matrices, or tensors, with a generative model tasked with producing the optimal tensor representation~\cite{wang2023automated} under specific geographic and social constraints. 
The work in~\cite{wang2020reimagining} frames urban planning as the generation of land-use configurations over graphs enriched with geographic attributes and contextual information. These configurations are encoded as multi-layered data representations representing Point of Interest (POI) distributions based on location and function. A graph autoencoder generates regional embeddings, which are fed into a GAN to assess generated plans using metrics such as willingness to pay, diversity of human activities, and social interaction. Despite its novelty, this approach struggles to model spatial hierarchies and exhibits training instability. In~\cite{wang2021deep}, land-use generation is conditioned on spatial context and textual planner guidance through a deep conditional variational autoencoder, yielding zone maps and POI layouts informed by contextual factors like geography and mobility. While effective in integrating planning expertise, it does not explicitly address hierarchical urban structures. 
To overcome these limitations,~\cite{wang2023human, wang2023automated} introduces a hierarchical conditional transformer that first produces zone-level functions before refining them into detailed grid-level plans. 
Similarly,~\cite{wang2023hierarchical} formulates planning as a two-tier hierarchical reinforcement learning (RL) problem, employing an actor-critic network and a Deep Q-Network to support decision-making at the zone and POI levels, respectively.
Likewise,~\cite{zheng2023spatial} models cities as graphs, utilizing GNN-based RL to make sequential planning decisions.

\subsection{The Underlying Generative Urban Planning Framework}
\vspace{12pt}
Generative urban planning formulates urban planing as conditional generation of land-use configurations, where an AI system learns to synthesize urban layouts that are optimized with respect to spatial, social, and policy-driven objectives.  This principled framework consists of two stages: learning representations of urban contexts and needs and generating land-use scenarios under those conditions.

\subsubsection{Representation: Modeling Urban Contexts and Human Needs}
\vspace{12pt}
The representation stage aims to encode the multifaceted conditions of a target urban area to plan into a structured embedding that guides the generative model. 
This involves integrating and representing diverse information: 
1) Geospatial Form: Captures the spatial arrangement and morphology of urban environments, including streets, buildings, parks, and infrastructure. These spatial patterns influence accessibility, land value, and neighborhood character. 
2) Human Mobility: Derived from data such as GPS traces, public transit logs, and mobile phone signals, mobility data reveals functional flows and commuting behaviors critical to understanding land-use needs. 
3) Social Interactions: Encompasses both physical interactions (e.g., in plazas, schools, offices) and digital engagement (e.g., social media activity). These patterns reflect community structure, cultural behavior, and public sentiment. 
4) Planner Requirements: Urban planners often express goals—such as increasing green space or housing density—via textual prompts, zoning rules, or categorical preferences. These must be encoded as flexible or strict constraints in the model.
To unify all these modalities, deep representation learning techniques such as autoencoders, convolutional neural networks (CNNs), and attention-based models are employed to transform high-dimensional inputs into semantically-rich, low-dimensional embeddings. These embeddings serve as a comprehensive representation of a region’s situational context and planning constraints, forming the condition for land-use generation.

\subsubsection{Generation: Producing Optimal Land-Use and Built Environment Configurations}
\vspace{12pt}
Given a learned representation, the generative stage synthesizes land-use plans using deep generative models, such as, Variational Autoencoders (VAEs)~\cite{kingma2019introduction}, Generative Adversarial Networks (GANs)~\cite{goodfellow2020generative,gulrajani2017improved,arjovsky2017wasserstein}, Autoregressive Models~\cite{kenton2019bert, NIPS2017_3f5ee243}, and other generative techniques. 
These generative urban planning models can be trained with a combination of machine learning objectives and planning-specific constraints, ensuring that generated designs remain both technically robust and aligned with urban planning principles.
Together, representation and generation form a pipeline where AI systems learn to "understand" the urban environment and "create" planning proposals that are not only technically valid but socially and contextually meaningful. 

\subsection{Embedding Generative Urban Planning to Real Systems: Flooding Resilience}
\vspace{12pt}
One application example is a redevelopment zone along the Gulf Coast, where legacy land-use patterns—such as dense residential clustering near flood-prone areas, fragmented green infrastructure, and spatially isolated critical facilities—have amplified vulnerability to storm surge and inland flooding. 
The central planning challenge is to reconfigure urban layouts to anticipate, absorb, and recover from flood events, while maintaining economic functionality and social continuity.
Generative urban planning offers a promising alternative by integrating AI with spatial design to generate land-use scenarios that are both realistic and adaptive to hydrological risks, offering planners new options beyond conventional rule-based methods.
Using multimodal data inputs, including flood simulation outputs, infrastructure risk maps, and planner-defined constraints, a conditional generative model can synthesize redevelopment layouts that elevate critical infrastructure, reposition vulnerable housing clusters, introduce green-blue corridors for water absorption, and optimize road networks for emergency access, all while complying with zoning and development guidelines. Such an approach aligns with the emerging "resilience-by-design" paradigm in urban planning \cite{meerow2016defining}.

\subsection{Limitations of Current AI Urban Planner Research}
\vspace{12pt}
Despite advances in AI for urban planning, current approaches face critical limitations across five foundational dimensions: theory, methodology, data, computing, and applications. 
1) Theory-wise, most AI models reduce planning to spatial optimization or land-use generation, thus, ignore core urban concepts such as zoning, resilience, social equity, and participatory governance, and often ignore the political and normative nature of planning decisions—undermining interpretability and legitimacy. 
2) Method-wise, models~\cite{wang2020reimagining,zheng2023spatial,wang2023hierarchical,wang2023human,wang2023automated,wang2021deep, wang2024dual} rely on single-step or loosely staged generation conditioned on static inputs that limit creativity and adaptivity, fail to capture multi-granularity dynamics, multiple planning objective trade-offs, and adaptive processes. 
3) Data-wise, existing city-specific datasets (e.g., Beijing~\cite{wang2021measuring}, NYC) lack clear evaluation criteria on scoring a good planning and a bad planning. 
4) Computation-wise, GenAI models are resource-intensive and prone to instability, especially in adversarial or variational settings; they lacks explainability and transparency due to the use of neural network based architectures, and often remain opaque to end users. 
5) Application-wise, existing generative planning methods ignore many domain expert developed tools. Be sure to notice that domain urban planners have developed various tools based on planning theory to assist each step of urban planning process, while existing pure neural networks based generation overlooks the power of these established practices.

\section{The Vision: Integrating VLM, Multimodal Reasoning, and Agentic AI for Urban Planning AI Agents}
\vspace{12pt}
To go beyond GenAI like VAEs, GANs, or diffusion models, we believe that Vision-Language Models (VLMs) integrate visual and textual modalities to produce outputs that are semantically meaningful and instruction-following~\cite{10.5555/3454287.3454289,liu2023visual}. This multimodal capacity allows AI systems to integrate maps, imagery, and policy goals to pursue complex planning objectives.
For example, urban design prompts (e.g., “generate a transit-oriented mixed-use layout”) can be aligned with satellite imagery, GIS data, and street-level visuals. This allows planners to interact with models using natural language, while ensuring that outputs remain spatially accurate and grounded in real-world data.
Agentic AI extends this capability by modeling the planning pipeline itself, with abilities for reasoning, task decomposition, and multi-step decision-making that mirror how human planners approach complex planning tasks. Agent-based urban planners can autonomously explore spatial trade-offs, simulate long-term effects, and iteratively refine plans based on feedback, environmental constraints, or guidance from human planners.
Together, VLMs and agentic systems enable an AI urban planner that is not only generative but also goal-driven and context-aware. Such AI systems can understand complex instructions, align outputs with multimodal data, and dynamically adapt proposals to evolving objectives.

\subsection{Agentic AI Abilities: Implications for Urban Planning}
\vspace{12pt}
Agentic AI introduces a new paradigm for urban planning by directly addressing its interdisciplinary and complex nature. Through the integration of reasoning, task decomposition, and the use of domain-specific tools, agentic AI has the potential to make planning processes more transparent, explainable, and practical.

\subsubsection{Reasoning}
\vspace{12pt}
Unlike static models that simply generate outputs, agentic AI can systematically analyze diverse objectives, constraints, and potential conflicts.
This reasoning ability enables the AI urban planners to balance objectives such as resilience, equity, and cost, while highlighting trade-offs and anticipating downstream consequences—making planning decisions more transparent and better aligned with real-world needs.

\subsubsection{Task decomposition}
\vspace{12pt}
Urban planning is too complex to be solved in a single step. Agentic AI addresses this challenge with a divide-and-conquer approach, decomposing the process into smaller, manageable tasks, each focused on a particular planning dimension such as land use, mobility, or infrastructure. These sub-tasks can follow their own objectives and constraints while remaining consistent with the overall plan, making the decision process more transparent, traceable, and easier for planners to engage with.

\subsubsection{Domain-specific tool usage}
\vspace{12pt}
Another strength of agentic AI is its ability to integrate practical tools developed by urban planning practitioners. These tools (e.g., simulation software, zoning analysis methods, resilience metrics) have been repeatedly validated in real-world projects and regulatory processes, and are grounded in decades of planning theory and professional practice. By leveraging such domain-expert tools, agentic AI can align its outputs with institutional standards and established planning principles rather than relying solely on static, data-driven inference. This makes the results more innovative, credible, and directly applicable to practice.
Through the integration of these abilities, agentic AI emerges as a practical paradigm for urban planning. It can handle complex and conflicting objectives in a transparent and traceable way, while aligning outputs with established planning standards and producing results that are credible to experts and accessible to stakeholders.

\begin{figure*}[!h] 
  \centering
  \includegraphics[width=\linewidth, trim = {0 2cm 0 0.5cm}]{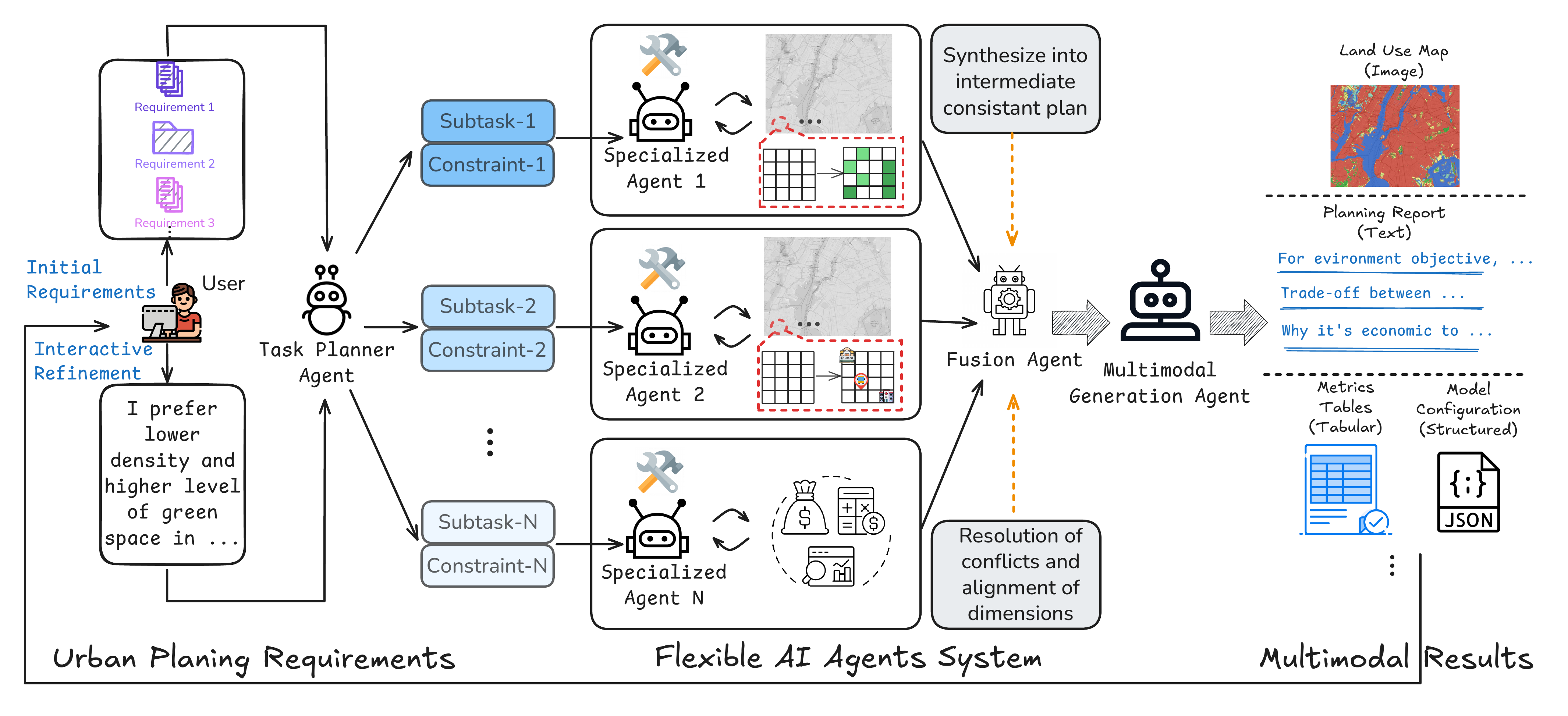} 
  \caption{Framework Overview of Agentic AI based Automated Urban Planner.}
  \label{fig:pipeline}
\end{figure*}

\subsection{The Agentic AI Urban Planner Framework}
\vspace{12pt}
Building on these capabilities, we propose an agentic AI urban planner framework that embeds AI into the planning workflow to produce clearer, more usable planning outputs. Figure~\ref{fig:pipeline} demonstrates the overall pipeline of an agentic AI urban planner.

\subsubsection{Task Planner Agent}
\vspace{12pt}
Given the diversity and potential conflicts among planning objectives and constraints, the agentic AI system starts with a Task Planner agent. 
The task planner agent has three primary functions. 
First, it builds a comprehensive view of the planning task by integrating diverse requirements and constraints, providing a transparent basis for scenario generation and evaluation. 
As the inputs span multiple modalities, such as textual documents, images, spatial data, and numerical indicators, the Task Planner interprets and integrates them into a coherent representation.
This multimodal fusion ensures a consistent and traceable understanding of context. It also makes explicit the relationships among objectives, constraints, and contextual factors, improving transparency.
Second, it breaks the workflow into smaller, manageable steps from the divide-and-conquer perspective.
Instead of generating a complete plan in a single step, it structures the work into sub-tasks that remain aligned with the overall goals. 
Each sub-task targets a specific planning dimension (e.g., land use, mobility, infrastructure) with its own objectives and constraints, while remaining consistent with the overarching goals.
Third, it defines and delegates sub-tasks by determining which Specialized Agents should take them on and which professional tools should be applied. 
This ensures that each task is matched to the right expertise, computational resources, and established planning practices.

\subsubsection{Specialized Agents}
\vspace{12pt}
Once the planning task has been decomposed, the resulting sub-tasks are assigned to Specialized Agents for implementation.
Each Specialized Agent is dedicated to a clearly defined scope of work, focusing on one particular dimension of the planning process. 
Instead of addressing the full complexity of the planning at once, they operate within the narrower boundaries of their designated sub-tasks, generating more accurate and reliable intermediate results.
These results enhance the transparency and traceability of the overall process by providing planners with interpretable outputs at each stage.
Moreover, Specialized Agents are empowered with domain-specific methods and practitioner-developed tools, enabling them to align outputs with professional expertise and regulatory standards.
Taken together, this division of labor creates a complementary set of agents that lays the foundation for integrating results into coherent planning alternatives.

\subsubsection{Fusion Agent}
\vspace{12pt}
After the Specialized Agents have completed their corresponding sub-tasks, their intermediate results are fed forward to the Fusion Agent. 
The Fusion Agent then aggregates these diverse outputs, resolve potential conflicts, and synthesize them into coherent planning alternatives.
In doing so, it ensures internal consistency across spatial, social, environmental, and economic dimensions that would otherwise remain fragmented.
Instead of simple aggregation, the Fusion Agent explicitly highlights trade-offs among competing objectives such as cost versus resilience or growth versus equity.
By reconciling differences across sub-task outputs, it generates planning candidates that are technically feasible, socially legitimate, and easier for stakeholders to evaluate and compare.
This agent is critical because, while Specialized Agents provide accurate results within their narrow scope, those results might be misaligned when considered together. 
The Fusion Agent addresses this challenge by acting as a system-level integrator, transforming partial results into consistent and actionable planning scenarios.

\subsubsection{Multimodal Generation Agent}
\vspace{12pt}
After the Fusion Agent synthesizes coherent planning candidates, the Multimodal Generation Agent eventually transforms then into multimodal outputs that are transparent, interpretable and accessible to different audiences.
Instead of limiting results to technical reports or abstract layouts, this agent produces complementary formats such as narrative summaries, tabular statistics, and spatial maps. 
These outputs make it easier for professional planners to evaluate trade-offs, while also enabling non-specialist stakeholders to understand and engage with the alternatives.
In contrast to conventional automated urban planning approaches that often formulate urban planning as a single optimization step, the agentic AI framework reconceptualizes urban planning as a multi-stage, reasoning-driven process, transforming the overall task from a black-box approach into an adaptive paradigm that is transparent, interpretable, and grounded in professional practice.
By organizing conflicting objectives into a structured problem set, the AI system provides urban planners with a transparent overview that clarifies priorities and guides subsequent steps.
Building on this foundation, specialized analyses across distinct dimensions draw on methods and tools established by domain experts to deliver credible and interpretable intermediate outputs. 
The synthesis of these intermediate results produces coherent alternatives that make trade-offs explicit and reinforce decision legitimacy, 
Finally, multimodal outputs ensure that results remain transparent to experts and accessible to broader stakeholders. 
This structured pipeline aligns closely with the routine workflow of urban planners, reinforcing clarity at the outset, credibility in intermediate analyses, and accessibility in final outputs. 
It enables planners to engage the system through natural-language prompts combined with diverse data inputs, iterating with the agents to test ideas and refine directions. 
By reasoning over complex contexts, the AI system can produce detailed textual narratives alongside preliminary spatial sketches, accelerating early-stage concept development while enriching it with greater depth. 
In doing so, agentic AI complements professional practice with outputs that are not only technically robust but also imaginative and accessible across multiple modalities.
To illustrate its practical value, we align the framework with the routine planning workflow, using flood resilience as a representative example to highlight how agentic AI augments planner-led processes at each stage.

\subsection{Flood Resilience Planning: A Motivating Example}
\vspace{12pt}
Flood resilience planning represents a classic challenge in urban planning because it concentrates many of the tensions that planners face in practice. 
Objectives are inherently conflicting: protecting property owners often requires costly defenses, while ensuring public safety may demand relocation or restrictions on development.
All of this must be managed within strict budget limits.
The tools and data required for decision-making are highly fragmented. Hydrological simulations, GIS-based vulnerability assessments, and economic cost–benefit models are usually run individually, making the integration difficult and time-consuming.
In traditional planning processes, key trade-offs are typically resolved through expert deliberation and then explained in technical documents, making it challenging for the public to understand or engage with.
These difficulties raise a pressing question: how can planners manage conflicting objectives, fragmented tools, and opaque decisions in a way that remains both rigorous and participatory? 
Agentic AI offers one possible answer. By reasoning through trade-offs, decomposing complex tasks, integrating expert methods, and presenting results in multimodal accessible forms, it can complement existing practice. 
To illustrate this potential, we trace how the agentic AI pipeline aligns with the familiar stages of the planning workflow, acting like a junior planner.

\subsection{Integrating Agentic AI into the Planning Workflow}
\vspace{12pt}
Urban planners typically follow a structured workflow that starts with problem scoping, followed by collaborative visioning, data collection and analysis, alternative generation, evaluation of trade-offs, and finally public deliberation and recommendation.
Each stage has distinct objectives, methods, and outputs, and together they form the backbone of day-to-day planning practice.
Building on the agentic AI framework introduced above, we now align its functions with these standard planning stages, showing how it can complement existing practice and address long-standing challenges such as fragmented tools, opaque trade-offs, and limited participation.

\subsubsection{Problem Scoping}
\vspace{12pt}
Problem scoping is the first step in the planning workflow, where planners define the scope of the challenge, clarify objectives, and identify key constraints. 
At this stage, they collect input from diverse stakeholders—residents, government agencies, and developers, while also aligning with policy documents, regulations, and budget limits. 
The goal is to establish a shared understanding of the priorities the planning process must address, from balancing flood protection with housing development to weighing costs against public safety. 
In practice, however, this step is often complicated by fragmented information, inconsistent stakeholder priorities, and limited transparency, since much of the discussion remains within expert groups.
Agentic AI can support planners at this stage by integrating diverse input materials such as flood maps, zoning rules, demographic data, and stakeholder goals into a single, coherent problem representation. 
Beyond simply compiling inputs, the Task Planner Agent highlights where objectives reinforce one another, where they conflict, and which constraints are likely to dominate later stages, establishing a structured problem set that identifies the objectives and constraints requiring attention in subsequent stages.
For human planners, this initial structured problem map provides a transparent baseline that clarifies priorities and reveals the tensions that must be carried forward, providing a clearer starting point for decomposing the task into manageable parts and moving into the collaborative visioning process.

\subsubsection{Collaborative Visioning}
\vspace{12pt}
Collaborative visioning follows problem scoping as the stage where planners and stakeholders work together to define common goals and imagine possible futures based on the shared understanding of the overall challenges and constraints. 
In this step, residents, officials, and developers discuss priorities for the community’s future, often drafting vision statements that capture aspirations for safety, livability, and economic growth.
In practice, collaborative visioning can be skewed by imbalances in participation, leading to situations in which dominant perspectives prevail and others are marginalized. 
As a result, disagreements are frequently left unresolved, which limits the transparency of outputs for subsequent stages.
Agentic AI addresses this challenge by analyzing inputs from meetings, documents, and surveys to identify both points of consensus and areas of disagreement, turning vague tensions into explicit trade-offs that enhance transparency and broaden opportunities for engagement.
In addition, the Task Planner Agent decomposes broad and often abstract aspirations into more concrete components. 
For instance, it can separate goals related to safety, livability, and economic growth into distinct components, each of which can be carried forward to specialized analyses. 
This decomposition not only clarifies priorities at the visioning stage but also ensures that subsequent evaluations remain transparent and traceable across specific dimensions. 
To support discussion, the Task Planner Agent can also organizes the decomposed goals into a small set of contrasting vision frames (e.g., safety-led, livability-led, growth-led), each with explicit assumptions and indicative targets, providing concrete reference points for dialogue without relying on abstract statements.
For planners, this yields a transparent baseline for visioning—one that highlights common ground, makes conflicts visible, and supplies structured materials (e.g., goal matrices and trade-off tables) to guide dialogue and participation.

\subsubsection{Data Intake and Analysis}
\vspace{12pt}
Following problem scoping and visioning, the planning process proceeds to data collection and analysis, where diverse technical tools are applied to build the evidence base for decision-making.
Hydrological simulations estimate flood risks, GIS overlays map vulnerable areas, economic models project costs and benefits, and demographic forecasts anticipate population trends. 
These analyses provide the technical foundation for developing alternatives. 
However, in practice they are often fragmented, as tools are applied independently, produce results in disparate formats, and lack mechanisms for integration across domains.
This fragmentation not only increases the workload for planners but also limits the transparency of how technical evidence is assembled and interpreted.
Agentic AI addresses these challenges through Specialized Agents, each dedicated to a specific dimension of the analysis.
A hydrology agent, for instance, can run flood models and standardize the results into formats that connect with infrastructure or economic assessments.
Similarly, other agents apply GIS methods, economic models, or demographic analyses, while ensuring that their outputs are interpretable and traceable.
Since these agents are designed to operate within defined scopes, they can deliver more accurate and consistent intermediate results, reducing duplication and making integration more straightforward.
For planners, this stage provides domain-specific results that are accurate, consistent, and traceable. 
Each output can be examined on its own, with assumptions and methods clearly documented. 
In this way, planners and stakeholders can verify the validity of each component, ensuring confidence in the evidence base before alternatives are developed.

\subsubsection{Alternative Generation}
\vspace{12pt}
After the data have been analyzed across different domains, the next stage is to generate planning alternatives.
In traditional processes, options are assembled from separate technical studies, which often yields inconsistencies across spatial, social, environmental, and economic dimensions.
Conflicts among objectives are not systematically articulated, and results lack a common structure for comparison, making rigorous evaluation difficult. 
Agentic AI strengthen this stage through the Fusion Agent, which synthesizes the standardized outputs from specialized analyses and synthesizes them into coherent planning candidates. 
By aligning assumptions and scales, the Fusion Agent creates a consistent basis for comparison across domains. 
It then evaluates each candidate against the stated constraints and makes trade-offs explicit, ensuring that planners can clearly assess the implications of different alternatives. 
For planners, the outcome is a not a disconnected set of reports but a structured set of comparable alternatives. 
Each option is presented with its assumptions, constraints, and quantified impacts, while also clarifying trade-offs among objectives. 
This combination provides a transparent basis for systematic evaluation and supports discussion not only among technical experts and decision-makers but also with the broader public.

\subsection{Evaluation \& Deliberation }
\vspace{12pt}
The final stage of the planning workflow is evaluation and deliberation, where proposed alternatives are assessed and discussed by both experts and the public. 
Traditionally, the findings are communicated through lengthy reports and technical documents. 
While these formats provide details, they are difficult for non-specialist audiences to interpret, limiting transparency and weakening opportunities for broad participation.
Agentic AI addresses this challenge through the Multimodal Generation Agent. 
Instead of relying on text-heavy documents, it translates analytical results into a range of outputs, including maps, tables, and narrative summaries. 
These multimodal representations allow experts to examine precise data while enabling broader stakeholders to engage with intuitive visualizations and plain-language descriptions.
For planners, this stage produces a set of planning deliverables that present each alternative through multiple, complementary perspectives.
Within these deliverables, trade-offs and assumptions are articulated systematically across alternatives, providing a clear and defensible basis for comparison.
The results are presented in varied formats such as narrative reports, tables, and maps, which allow experts to examine detailed evidence and enables policymakers and community members to engage with accessible representations.
As a result, planners are able to evaluate alternatives in a process that is more transparent, inclusive, and directly connected to decision-making. 
This example demonstrates that agentic AI can be seamlessly integrated into the standard planning process of urban planners, aligning with each stage of practice rather than replacing it. 
Like a junior planner, it contributes constructively throughout the workflow by clarifying objectives at the outset, structuring aspirations into manageable components, providing consistent and credible analyses across domains, assembling alternatives into coherent options, and presenting results in accessible forms that support deliberation.
Through these capabilities, the agentic AI system enables planners to navigate complexity with greater clarity, evaluate trade-offs with stronger evidence, and communicate outcomes in ways that are more transparent, participatory, and inclusive.

\section{What to Consider in Learning Urban Planning AI Agents}
\vspace{10pt}
\subsection{Detecting and Prioritizing Real Human Needs Before Urban Planning}
\vspace{12pt}
Urban planning must be grounded by real human needs including:
\begin{itemize}
    \item Essential needs: safe and affordable housing, access to clean water, sanitation, healthcare, and food; 
    \item Functional needs: mobility infrastructure, public transport, education, and employment access; 
    \item Resilience needs: climate adaptation (e.g., heat mitigation, flood protection), risk reduction, and emergency response capacity; 
    \item Aspirational needs: access to green space, social connection, cultural expression, and digital inclusion. 
\end{itemize}

Our perspective is to leverage AI to sense and infer urban planning needs from both explicit signals (e.g., surveys, civic platforms) and implicit behaviors (e.g., movement traces, online discourse, built environment scans). For example, by mining public discourse from social media, 311 complaints, and community forums, we can exploit NLP and LLMs to extract topics (e.g., transit dissatisfaction, service breakdowns) and sentiments that reflect resident pain points, detect urgency, importance, locations in user-generated contents, mapping emotional intensity to geographic locations. 
As another example, by analyzing street imagery, drone footage, or satellite data, we can detect missing sidewalks, poor lighting, slum expansion, or playground deficits, and classify urban elements relevant to walkability, safety, or accessibility.
The third example is: by modeling GPS traces, call detail records, or smartcard data, AI can identify mismatches between population flows and service coverage, such as last-mile connectivity gaps, transit deserts, or areas with extreme commuting burdens.

Besides, Agentic AI can be leveraged to simulate diverse community behaviors under planning scenarios and infer emergent needs by observing agent outcomes (e.g., failure to access services, congestion points, or repeated exposure to risk). 

Finally, geospatial foundation models, pretrained on large-scale maps, satellite images, and terrain data, can infer land-use, infrastructure density, and vulnerability, providing transferrable knowledge to tackle the issues of under-mapped or data-scarce regions.

\subsection{Differentiate GenAI in Strategic Macro Planning and Scenario Micro Planning}
\vspace{12pt}
Generative urban planning should consider the distinction between generic global planning and scenario-specific local planning. 
Firstly, strategic planning, also known as, macro-scale land-use planning, focuses on long-term, citywide or regional spatial configurations, aiming to optimize zoning, connectivity, and land allocation over 10–30 year horizons. 
Secondly, scenario planning, often referred to as tactical urban design or micro-planning, targets specific neighborhoods or blocks, adapting to contextual challenges such as flooding, extreme heat, or post-disaster recovery. 

Strategic planning aims to capture self-organized planning objectives, such as diversity over mixed land uses, connectivity of urban functions, accessibility to transportation, healthcare, utilities, job opportunities,  and simulation to verify the optimal planning objectives over long-term dynamics. 
On the other hand, scenario planning aims to prioritize conditional generation under fine-grained constraints. 
GenAI models, such as hierarchical VAEs, conditional diffusion models, and scenario-based reinforcement fine-tuning can empower guided generation of urban planning.

\subsection{Human-Machine Collaborative Planning}
\vspace{12pt}
Urban CoDesign (i.e., human-machine collaborative planning) enabled by conversational generative AI is a paradigm where urban planners and AI systems co-design planning solutions through natural language and interactive feedback. 
CoDesign highlights that AI is a responsive partner that generates, explains, and adapts plans based on iterative human input, instead of generating fixed outputs. This is important because urban planning involves complex trade-offs, contextual sensitivity, and normative goals that pure data-driven models often overlook. Feedback from human planners can ground generative planning in policy, social, economic, cultural dimensions. 

Possible solutions to achieve the codesign  are: integration of  multimodal generative planners, instruction-tuned vision language models, and interactive machine learning that respond to planners’ queries, critiques, and constraints human-machine conversations. Imagine a planner might say, “Propose a higher-density zoning plan that avoids flood zones and preserves public parks,” and the system would iteratively refine the layout while visualizing trade-offs. 

\subsection{Leveraging Digital Twins For A Close Simulation, Measurement, Planning Loop}
\vspace{12pt}
Another direction is to leverage digital twins to develop a simulation–measurement–generation loop of urban planning. Traditional urban planning suffers from long feedback cycle, often taking years to observe the social and economic impacts of spatial interventions. 

As high-fidelity, real-time virtual replicas of urban environments, digital twins can simulate how spatial configuration plannings affect economic, mobility, and social dynamics. These simulations generate synthetic yet actionable data that feed into measurement models to quantify key urban indicators (e.g.,  sustainability, accessibility, vibrancy, safety, resilience, perceived well-being). 
These urban indicators then serve as feedback signals for learning neural urban planning policy networks that guide generative planning toward optimal configurations under evolving urban goals. 
Such design not only accelerates the evaluation of planning alternatives, but also supports adaptive planning where planning decisions are informed by iterative simulation and measurable impact assessment within digital twins.

\bibliographystyle{IEEEtran}
\bibliography{main}

\end{document}